# A Comparative study of Transportation Problem under Probabilistic and Fuzzy Uncertainties

Arindam Chaudhuri*
Lecturer (Mathematics and Computer Science)
Meghnad Saha Institute of Technology
Nazirabad, Uchchepota, Kolkata, India
Email: arindam_chau@yahoo.co.in
*Corresponding Author

Dr. Kajal De
Professor of Mathematics,
School of Science,
Netaji Subhas Open University,
Kolkata, India

**Abstract**

Transportation Problem is an important aspect which has been widely studied in Operations Research domain. It has been studied to simulate different real life problems. In particular, application of this Problem in NP- Hard Problems has a remarkable significance. In this Paper, we present a comparative study of Transportation Problem through Probabilistic and Fuzzy Uncertainties. Fuzzy Logic is a computational paradigm that generalizes classical two-valued logic for reasoning under uncertainty. In order to achieve this, the notation of membership in a set needs to become a matter of degree. By doing this we accomplish two things viz., (i) ease of describing human knowledge involving vague concepts and (ii) enhanced ability to develop cost-effective solution to real-world problem. The multi-valued nature of Fuzzy Sets allows handling uncertain and vague information. It is a model-less approach and a clever disguise of Probability Theory. We give comparative simulation results of both approaches and discuss the Computational Complexity. To the best of our knowledge, this is the first work on comparative study of Transportation Problem using Probabilistic and Fuzzy Uncertainties.

## 1. Introduction

The transportation problem is a special category of Linear Programming Problem [5], [7]. It has been widely studied in Logistics and Operations Management where distribution of goods and commodities from sources to destinations is an important issue. The task of distributor's decisions can be optimized by reformulating the Distribution Problem as generalization of the classical Transportation Problem [5], [7]. The conventional Transportation Problem can be represented as a mathematical structure which comprises an Objective Function subject to certain Constraints. In classical approach, transporting costs from $M$ sources or wholesalers to $N$ destinations or consumers are to be minimized. It is an Optimization Problem which has been applied to solve various NP-Hard Problems.

The origin of transportation methods dates back to 1941 when F. L. Hitchcock presented a study entitled *The Distribution of a Product from Several Sources to Numerous Localities*. This presentation is considered to be the first important contribution to the solution of Transportation

Problems. In 1947 T. C. Koopmans presented an independent study called *Optimum Utilization of the Transportation System*. These two contributions helped in the development of transportation methods which involve a number of shipping sources and a number of destinations. Within a given time period each shipping source has a certain capacity and each destination has certain requirements with a given cost of shipping from source to destination. The Objective Function is to minimize total transportation costs and satisfy destination requirements within source requirements [5], [7]. However, in real life situations, the information available is of imprecise nature and there is an inherent degree of vagueness or uncertainty present in the problem under consideration. In order to tackle this uncertainty the concept of Fuzzy Sets [12] can be used as an important decision making tool. Imprecision here is meant in the sense of vagueness rather than the lack of knowledge about parameters present in the system. Fuzzy Set Theory thus provides a strict mathematical framework in which vague conceptual phenomena can be precisely and rigorously studied.

In this work, we have concentrated on the comparative study of Transportation Problem under Probabilistic and Fuzzy Uncertainties [1], [9]. The proposed approach allows us to accomplish direct Fuzzy extension of classical numerical Simplex Method [2]. We compare the results obtained using Fuzzy and Probabilistic approaches. The simple special method for transformation of Frequency Distributions into Fuzzy Numbers without lost of useful information is used to achieve the comparability of uncertain initial data in fuzzy and random cases. The Fuzzy interval is achieved using Trapezoidal Fuzzy Numbers [12]. This Paper is organized as follows. In section 2, we briefly discuss about Transportation Problem. In next section we define Trapezoidal Membership Function. This is followed by some work related to Transportation Problem. In section 5, we give the formulation of Transportation Problem under Probabilistic and Fuzzy Uncertainties. This formulation is followed by a Numerical Example. Finally, in section 7 conclusions are given.

**2. Transportation Problem**

The general Transportation Problem is concerned with distributing any commodity from any group of supply centers called sources to any group of receiving centers called destinations in such a way as to minimize the total distribution cost. We now give the mathematical formulation of Transportation Problem [5], [7].

Let us consider $M$ locations (origins) as $O_1,........,O_m$ and $N$ locations (destination) as $D_1,........,D_n$ respectively. Let $a_i \geq 0; i = 1,........,M$ the amount available at $i^{th}$ plant $O_i$. Let the amount required at $j^{th}$ shop $D_j$ be $b_j \geq 0; j = 1,........,N$. Further, the cost of transporting one unit of commodity form $i^{th}$ origin to $j^{th}$ destination is $C_{ij}; i = 1,........,M, j = 1,........,N$. If $x_{ij} \geq 0$ be the amount of commodity to be transported from $i^{th}$ origin to $j^{th}$ destination, then the problem is to determine $x_{ij}$ so as to Minimize,

$$z = \sum_{i=1}^{M} \sum_{j=1}^{N} x_{ij} c_{ij} \quad (1)$$

subject to constraints:

$$\sum_{j=1}^{N} x_{ij} = a_i; i = 1,........,M \quad (2) \qquad \sum_{i=1}^{M} x_{ij} = b_j; j = 1,........,N \quad (3)$$

$$x_{ij} \geq 0, \forall i, j \quad (4)$$

This Linear Programming Problem is called Transportation Problem. The necessary and sufficient condition for the existence of feasible solution to Transportation Problem is:

$$\sum_{i=1}^{M} a_i = \sum_{j=1}^{N} b_j \quad (5)$$

The constraints $\sum_{j=1}^{N} x_{ij} = a_i$ and $\sum_{i=1}^{M} x_{ij} = b_j$ represent $M+N$ equations in $MN$ non-negative variables. Each variable $x_{ij}$ appears in exactly two constraints, one is associated with origin and other with destination. If we put the data in matrix from as shown in Figure 1, the elements of matrix are either 0 or 1. In a Transportation Table, an ordered set of four or more cells is said to form a loop if [5], [7]:

1. Any two adjacent cells in the ordered set lie in same row or in same column.
2. Any three or more adjacent cells in the ordered set do not lie in same row or in same column.

A feasible solution to Transportation Problem is Basic if and only if the corresponding cells in Transportation Table do not contain a loop. The initial basic feasible solution is obtained by applying [5], [7]: (i) North-West Corner Rule and (ii) Vogel`s Approximation Method.

|  | **DESTINATIONS** | | | | |
|---|---|---|---|---|---|
|  | **D₁** | **D₂** | ------- | **Dₙ** | |
| **O₁** | C₁₁ | C₁₂ | ------- | C₁ₙ | **a₁** |
| **O₂** | C₂₁ | C₂₂ | ------- | C₂ₙ | **a₂** |
| ⋮ | ------- | ------- | ------- | ------- | ⋮ |
| **Oₘ** | Cₘ₁ | Cₘ₂ | ------- | Cₘₙ | **aₘ** |
|  | **B₁** | **B₂** | ------- | **bₙ** | |

(ORIGINS labels the rows)

**Figure 1: Transportation Problem**

## 3. Trapezoidal Membership Function

A trapezoidal membership function is defined by four parameters viz., $a, b, c, d$ as follows [12]:

$$trapezoid(x;a,b,c,d) = \begin{cases} 0, & x \leq a \\ \frac{x-a}{b-a}, & a \leq x \leq b \\ 1, & b \leq x \leq c \\ \frac{d-x}{d-c}, & c \leq x \leq d \\ 0, & d \leq x \end{cases} ; x \in \Re \quad (6)$$

## 4. Related Work

The Transportation Problem is one of the most important and most studied Problems in Operations Management domain [5], [7]. Much of the work on Transportation Problem is motivated by real life applications. Indeed, the numerous applications of Transportation Problem bring life to research area and help to direct future work. The first references to Transportation Problem dates back to 17$^{th}$ Century. The problem has been studied intensively in Optimization Techniques since then. In Mathematics and Economics, Transportation Theory is the name given to the study of optimal transportation and allocation of resources. The problem was initially formalized by French mathematician Gaspard Monge in 1781. Major advances were made in the field during World War II by Russian Mathematician and Economist Leonid Kantorovich. Consequently, the problem as it is now stated is sometimes known as Monge-Kantorovich Transportation Problem.

In 1979, Isermann [6] introduced algorithm for solving this problem, which provides effective solutions. The Ringuest and Rinks [10] proposed two iterative algorithms for solving linear, multi-criteria Transportation Problem. Similar solution was proposed by Bit [5], [7]. The different effective algorithms were worked out for Transportation Problem, but with regarding parameters of task described in form of Real Numbers. Nevertheless, such conditions are fulfilled seldom or almost never because of on natural uncertainties we meet in real world problems. For example, it is hard to define stable cost of specific route. In the work by Das, this problem was solved in case of Interval uncertainty of transporting costs. In works by Chanas and Kuchta [3], [4], the approach based on Interval and Fuzzy coefficients had been elaborated. The further development of this approach introduced in work of Waiel [11]. All the above mentioned works introduce restrictions in form of Membership Function. This allows transformation of initial Fuzzy Linear Programming Problem into the usual Linear Programming Problem by use of well defined analytic procedures [12]. However, in practice Membership Functions, which describes uncertain parameters of used models, can have considerable complicated forms. In such cases, the numerical approach is needed. The main technical problem when constructing numerical Fuzzy Optimization algorithm is to compare fuzzy values. To solve this problem we use the approach based on $\alpha$-level representation of Fuzzy Numbers and Probability Estimation of the fact that given interval is greater than or equal to other interval [1], [3], [4], [8]. The Probabilistic approach was used only to infer set of formulae for deterministic quantitative estimation of intervals inequality or equality. The method allows comparison of interval and real number and to take into account implicitly widths of intervals ordered.

## 5. Transportation Problem under Probabilistic and Fuzzy Uncertainties

In this work we minimize the costs in Transportation Problem and maximize distributor's profits under the same conditions. Here, the distributor deals with $M$ wholesalers and $N$ consumers. Let $a_i, i = 1,......., M$ be the maximal quantities of goods that can be supplied by wholesalers and $b_j \geq 0, j = 1,......., N$ be the maximal good requirements of consumers. In accordance with signed contracts distributor must buy at least $p_i$ units of goods at price of $t_i$ monetary units from each $i^{th}$ wholesaler and to sell at least $q_j$ units of goods at price of $s_j$ monetary units to each $j^{th}$ consumer. The total transportation cost of delivering good units from $i^{th}$ wholesaler to $j^{th}$ consumer is denoted as $c_{ij}$. There is reduction in prices $k_i$ for distributors if they purchase greater quantity of goods than stipulated quantity $p_i$ and also reduced prices $r_j$ for consumers if they purchase greater quantity than contracted quantity $q_j$. The problem is to find optimal good quantities $x_{ij}$ $(i = 1,........., M; j = 1,......., N)$ delivering from $i^{th}$ wholesaler to $j^{th}$ consumer maximizing distributor's total benefit $D$ under restrictions. We assume that all above mentioned parameters are fuzzy ones and the resulting optimization problem can be formulated as [9], [12]:

$$\max D = \sum_{i=1}^{M}\sum_{j=1}^{N}(z_{ij} * x_{ij}) \quad (7)$$

subject to constraints:

$$\sum_{j=1}^{N} x_{ij} \leq a_i, i = 1,........., M \quad (8) \qquad \sum_{i=1}^{M} x_{ij} \leq b_j, j = 1,........., N \quad (9)$$

$$\sum_{j=1}^{N} x_{ij} \geq p_i, i = 1,........., M \quad (10) \qquad \sum_{i=1}^{M} x_{ij} \geq q_j, j = 1,........., N \quad (11)$$

where, $z_{ij} = r_j - k_i - c_{ij}; i = 1,........., M; j = 1,........., N$ and $D$, $z_{ij}$, $a$, $b$, $p$, $q$ are fuzzy values.

To decide the problem specified by Equations (7) to (11), the numerical method based on α-cut representation of fuzzy numbers and probabilistic approach to interval and fuzzy interval comparison has been discussed. We make use of direct fuzzy extension of Simplex Method. To estimate effectiveness aspect, the results of Fuzzy Optimization [2], [12] were compared with those obtained from Equations (7) to (11) when all uncertain parameters were considered as random values which are normally distributed and parameters were considered as Real Numbers. Generally, we have a problem with different precisions of representation of uncertain data. For instance, one part of parameters used can be represented through Trapezoidal Fuzzy Numbers form on basis of expert's opinions and at the same time, other part of them can have the form of Histogram Frequency Distributions of considerable complicated form obtained from Statistical analyses. In these cases, the correct approach is to transform all uncertain data available to the form of smallest certain level. Thus, we transform data represented in form of Frequency Distributions or Histogram to Membership Functions of Fuzzy Numbers [1], [3], [9]. To present initial data in Fuzzy Number form, we apply a technique, which develops Membership Function on basis of Frequency Distribution, if such exists, or directly using Histogram. In the simplest case of Normal Frequency Distributions [5], [7], they can be exhaustively described be their averages $m$ and standard deviations $\sigma$. In more complicated situations it seems better to use Histograms. Thus we use a numerical technique which allows us to transform Frequency Distribution or Histogram to Trapezoidal Fuzzy Number [12]. As an illustration, let us consider reduction of Frequency Distribution to Fuzzy Number.

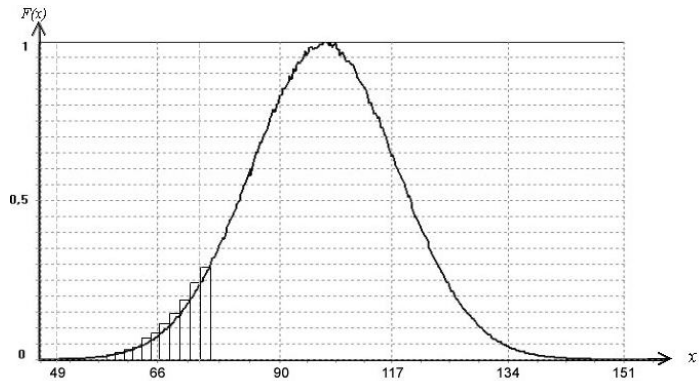

**Figure 2: Frequency distributions to be transformed**

We follow the following steps of algorithm [1], [3], [4], [8], [9]:

**Step 1:** In the interval within smallest value $x_{min}$ ($x_{min} = 50$) and maximum value $x_{max}$ ($x_{max} = 151$), we define function $F(x_i)$ is defined as surface area under curve (in Figure 2) from $x_{min}$ to current value of $x_i$. As a result, we get the cumulative function is obtained as shown in Figure 3. It is easy to see that function $F(x_i)$ is actually the probability of $x < x_i$.

**Step 2:** Using obtained Cumulative Function $F(x)$ we give four decision values of $F(x_i)$ at $i = 0, 1, 2, 3$, which define mapping of $F(x)$ on $x$ in such a way that they provide upper and bottom α-levels of Trapezoidal Fuzzy Number. In the example given in Figure 3, intervals $[95,105]$ and $[78,120]$ are in essence the 30% and 90% of probability confidence intervals. As a result we get Trapezoidal Fuzzy Interval represented by the quadruple $[78,95,105,120]$.

It is easy to see that transformation accuracy depends only on suitability and correctness of upper and bottom confidence intervals chosen. It is worthy to note that the main advantage of this method is that it can be successfully used in both cases viz., when we have initial data in a form of Frequency Distribution Function and a rough Histogram form. The method allows us to represent all uncertain data in uniform way as Trapezoidal Fuzzy Intervals [1], [3], [4], [8], [9]. The solution for the method represented by Fuzzy Programming Problem in Equations (7) to (11) is realized by performing all Fuzzy Numbers as sets of α-cuts. In fact, it reduces Fuzzy Problem into the set of Crisp Interval Optimization Problem. The final solution has been obtained numerically using Probabilistic approach to Interval comparison. The Interval arithmetic rules needed were realized with a help of Object-Oriented Programming tools. The standard Monte Carlo procedure [5], [7] was used for realization of Probabilistic approach to the description of uncertain parameters of Optimization Problem described by Equations (7) to (11). For each randomly selected set of real valued parameters of Optimization Problem we solve using Linear Programming Problem [5], [7], [12].

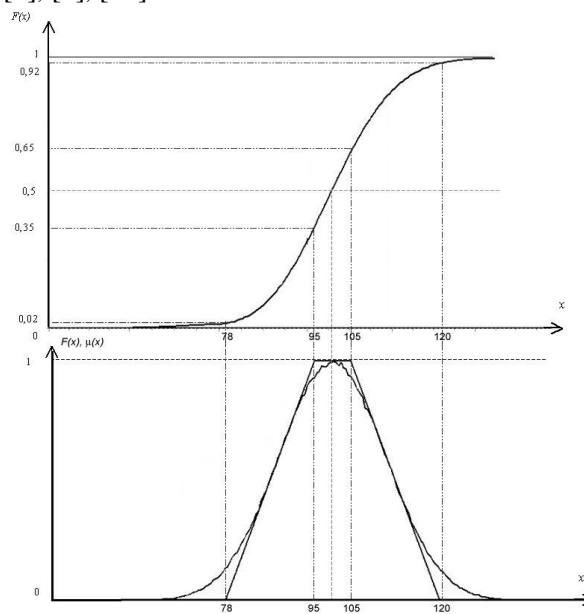

**Figure 3: The transformation of Cumulative Function to a Fuzzy Number**

## 6. Numerical Example

In this section, we present a Numerical Example for Transportation Problem using Probabilistic and Fuzzy Uncertainties. To compare the results of Fuzzy Programming with those obtained when using Monte Carlo Method [8], [9] all uncertain parameters were previously resolved using Gaussian Frequency Distributions. The averages of them are presented in Table 1. For simplicity we take all standard deviation values $\sigma = 10$.

| $a_1 = 460$ | $b_1 = 410$ | $p_1 = 440$ | $Q_1 = 390$ | $t_1 = 600$ | $s_1 = 1000$ | $K_1 = 590$ | $r_1 = 990$ |
|---|---|---|---|---|---|---|---|
| $a_2 = 460$ | $b_2 = 510$ | $p_2 = 440$ | $Q_2 = 490$ | $t_2 = 491$ | $s_2 = 1130$ | $K_2 = 480$ | $r_2 = 1100$ |
| $a_3 = 610$ | $b_3 = 610$ | $p_3 = 590$ | $q_3 = 590$ | $t_3 = 581$ | $s_3 = 1197$ | $k_3 = 570$ | $r_3 = 1180$ |
| $c_{11} = 100$ | $c_{12} = 30$ | $c_{13} = 100$ | | | | | |
| $c_{21} = 110$ | $c_{22} = 36$ | $c_{23} = 405$ | | | | | |
| $c_{31} = 120$ | $c_{32} = 148$ | $c_{33} = 11$ | | | | | |

**Table 1: Average values of Gaussian distributions of uncertain parameters**

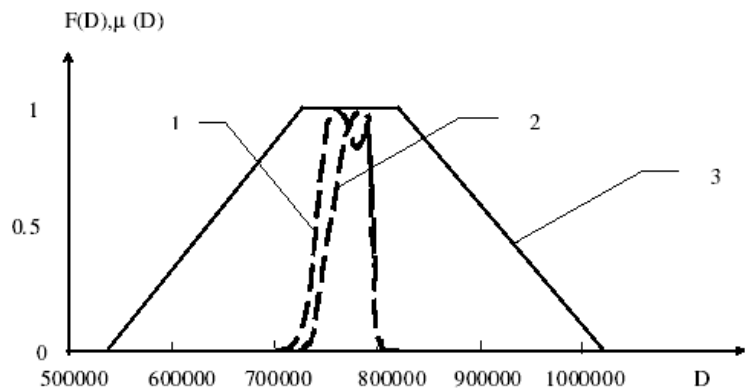

Fig. 8 Frequency distribution F and fuzzy number µ for optimized benefit D:
1 - Monte-Carlo metod for 10 000 random steps;
2 - Monte-Carlo metod for 100 000 000 random steps;
3 - Fuzzy approach

The results we obtained using Fuzzy Optimization Method and Monte Carlo Method [8], [9], [12] i.e., Linear Programming with real valued but random parameters are given in Figures 4 to 8, considering $M = N = 3$, where final frequency distributions $F$ are drawn by dotted lines, Fuzzy Numbers $\mu$ are drawn by continuous lines. It is easy to see that Fuzzy approach give us some more wide Fuzzy Intervals [3], [4], [9] than Monte Carlo Method. Using Probabilistic Method we get two-extreme results whereas Fuzzy approach always gives us results without ambiguity. It is worth considering that Probabilistic Method demands too many random steps to obtain smooth Frequency Distribution of resulting benefit $D$. Thus, this method is not used in practice.

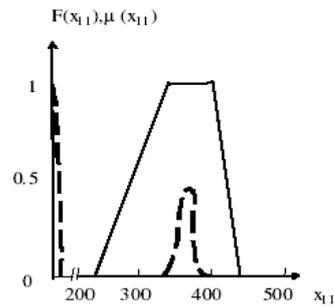
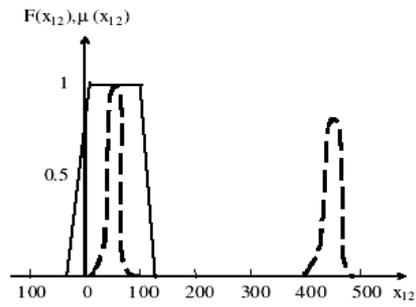

Fig. 4 Frequency distribution $F$ and fuzzy number $\mu$ for optimized $x_{11}$

Fig. 5 Frequency distribution $F$ and fuzzy number $\mu$ for optimized $x_{12}$

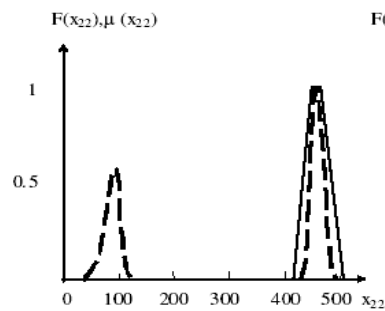
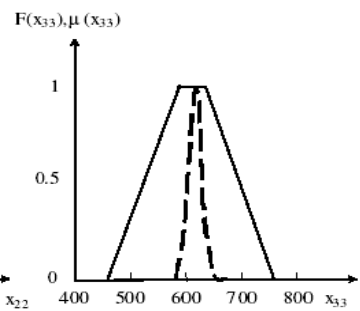

Fig. 6 Frequency distribution $F$ and fuzzy number $\mu$ for optimized $x_{22}$

Fig. 7 Frequency distribution $F$ and fuzzy number $\mu$ for optimized $x_{33}$

## 7. Conclusion

The Transportation Problem is considered in this paper using Probabilistic and Fuzzy uncertainties. The Transportation Problem is solved using Fuzzy technique is based on $\alpha$-level representation of Fuzzy Numbers and Probability Estimation of the fact that given interval is greater than or equal to other interval. The proposed method makes it possible to extend Simplex Method using Fuzzy Numbers. The numerical results obtained using Fuzzy Optimization Method and Monte Carlo Method with Linear Programming using real valued and random parameters show that Fuzzy approach has considerable advantages in comparison with Monte Carlo Method especially from computational point of view. The results obtained can be considerably improved by using a hybridization of Neuro-Fuzzy approach.

**Acknowledgement**

We wish to thank all anonymous referees for their many useful comments and specially Prof. Dipak Chatterjee, Department of Mathematics, St. Xavier's College, Calcutta for his constant encouragement and support in preparing this Paper.